\documentclass[11pt]{article}

\usepackage[final]{acl}
\usepackage[colorinlistoftodos]{todonotes}
\usepackage{times}
\usepackage{latexsym}
\usepackage{amsmath}
\usepackage[T1]{fontenc}
\usepackage{enumitem}
\setlist[itemize]{align=parleft,left=0pt..1em}
\usepackage[utf8]{inputenc}

\usepackage{microtype}

\usepackage{inconsolata}

\usepackage{graphicx}

\title{Geometry of Semantic Space:\\
Comparative Study of Discrete and Continuous Models}

\author{
Gabriel Bounias\textsuperscript{1,2} \quad
Sabine Ploux\textsuperscript{2}
\\[0.5em]
\textsuperscript{1} ISC-PIF (Institut des Systemes Complexes de Paris IdF), CNRS, France
\\
\textsuperscript{2}
CAMS (Centre d’analyse et de mathématique sociales), CNRS \& EHESS, Paris, France
\\[0.5em]
{\small \texttt{gabriel.bounias@ens-lyon.fr}}
\qquad
{\small \texttt{sabine.ploux@ehess.fr}}
}

\begin{document}

\maketitle

\begin{abstract}
This work examines the semantic geometry underlying NLP models. We compare supervised vector embeddings, such as CamemBERT, with lexical co-occurrence graphs that encode semantic relations more directly. While transformer-based embeddings achieve strong performance, their induced geometries often display unsatisfactory distributions. In contrast, graph-based models reveal a clearer and more human-readable organization of meaning. 

We have implemented a methodology that allows us to perform a comparative analysis either  
based on the structure of the graphs or based on the topology of the embeddings induced  by these two approaches.

The results of the comparison -- applied to the French "Great National Debate" corpus  a collection of citizen contributions to the public debate-- show a similar local topology but a very different overall structure and topology.
Theses findings suggest  complementary perspectives between deep supervised models and graph-based models, considering a new pathway to guide neural architectures toward more stable and interpretable convergence with graphs structures.

\end{abstract}
\section{Introduction}

Natural languages structure a space of meaning where each unit finds its place based on semantic proximity. This emerging geometry is an essential marker of how an NLP model reflects the semantic functioning of the language it deals with. This production of internal geometry reflects a more or less effective understanding of the use of linguistic units.

In the meaning space representation word, there are two main approaches. Vector models derived from deep learning (BERT, GPT, etc.) map each word into a large-dimensional continuous space. Conversely, graph models rely on discrete structures, where linguistic entities are linked by co-occurrence or lexical proximity. This construction makes the emergence of neighborhoods more explicit and allows for direct geometric analysis of meaning relationships. 

We therefore hypothesize that the quality of an NLP model can be evaluated through the geometric consistency of the structures it induces. Comparing the geometries derived from vector and graph models, applied to the same corpus, highlights their respective limitations, but also their complementarity. Our goal is to analyze how these structures reflect, or conversely distort, the organization of meaning in language, as conceived by humans.

\section{Modelling}

\subsection{Corpus and linguistic unit} 

\quad In this comparative study, we must first agree on a common constructive basis. To do this, we need to base the construction of models on the analysis of the same corpus. The analysis is based on the Grand Débat National corpus, derived from the 2019 public consultation that followed the Gilets Jaunes movement. This corpus, comprising around 10 million sentences, provides sufficient lexical and thematic diversity to explore the dynamics of meaning within a well-defined discursive context. This choice helps limit interpretative complexity while ensuring comparability between the resulting semantic geometries. Both models — the vector-based and the graph-based — are applied to the same textual foundation, following a lemmatization step (using \emph{Simplemma}, \cite{barbaresi2023simplemma}) to neutralize morphological variations and ensure linguistic consistency throughout the analysis. 

The comparative analysis also requires defining a common semantic unit to ensure the coherence of comparisons. Rather than working with isolated words prone to contextual polysemy, this study adopts regular co-occurrence cliques as the linguistic unit \cite{ji2003lexical}. A co-occurrence clique is defined as a set of words that appear together with a higher-than-expected probability within the same context, like a sentence skeleton. This approach captures polysemy more precisely, as each clique represents a contextualized and recurrent configuration of meaning within the corpus. Co-occurrences are measured using Pointwise Mutual Information (PMI), which quantifies the strength of association between two words beyond chance. By applying thresholds on frequency and PMI, a graph connecting highly correlated words is constructed, from which maximal cliques are extracted to serve as reference semantic units. This methodological choice provides strong semantic precision while ensuring a consistent foundation for comparing the geometry of meaning emerging from both paradigms. We can now begin building the two models on this basis. 
%On this basis, we can now begin building both types of models. (?)

\subsection{Graph model}

In constructing the graph-based model, the first step is to define a simple yet robust measure to quantify the proximity between cliques of co-occurrences, i.e., the semantic units extracted from the corpus. We introduce a one-parameter similarity function, noted $s$, which combines two complementary dimensions of linguistic proximity: on the one hand, contextual similarity, captured by the regular co-occurrence of words within similar contexts, and on the other, lexical overlap, captured by shared words between cliques. Formally, the similarity between two cliques $C_1$ and $C_2$ is defined as:

\begin{align*}
\label{eq:simil}
s(C_1,C_2)=\Bigg(1+\lambda\frac{|C_1\cap C_2|}{\max(|C_1|,|C_2|)}\Bigg) \times &\\
\sum_{(w_1,w_2)\in C_1\times C_2}\frac{\operatorname{PPMI}(w_1,w_2)}{|C_1||C_2|}&
\end{align*}

This equation simultaneously captures thematic proximity through the cross-average of Positive PMI values and lexical overlap, via the left-hand coefficient weighted by $\lambda$. The parameter $\lambda$ controls the balance between the contextual and lexical components, ensuring that both contribute comparably to the final similarity measure.

Once this similarity is computed between all pairs of cliques, we construct a graph $G_c$ whose nodes correspond to the co-occurrence cliques, and whose edges connect cliques that exceed a similarity threshold on $s$.

\subsection{Supervised vector model}

\subsection{Construction of the continuous model}

\quad To model meaning within a continuous semantic space, we rely on CamemBERT \cite{martin2020camembert}, a RoBERTa-based \cite{liu2019roberta}  Transformer model pre-trained on 138 GB of French text from the OSCAR corpus using the Masked Language Modeling (MLM) objective. CamemBERT provides a bidirectional encoder architecture, where each token attends to all others within the same sequence. The model operates with $12$ attention layers and $12$ heads per layer, producing contextualized embeddings of dimension $768$. Its tokenizer, SentencePiece, segments words into subword units to better capture morphological and lexical variations specific to French. This setup ensures that the model handles polysemy and inflectional morphology in a robust and linguistically coherent way.

\quad We use this model to embed the co-occurrence cliques extracted from the \textit{Grand Débat National} corpus\footnote{\url{https://www.data.gouv.fr/datasets/grand-debat-national-propositions}}. Each clique $C$ is represented by averaging the contextual embeddings of its constituent words, computed across $n_p$ sentences in which the clique occurs most prominently. And each resulting vector $\mathbf{v}_C$ take place in an high-dimensional continuous semantic space, where each point corresponds to a meaning.

An useful remark is that this representation enables the derivation of a graph structure $G_b$ based on cosine or Euclidean similarity between clique embeddings. Finally we made procedure to construct 2 types of graphs $G_c = (V,E_c,W_c)$ and $G_b = (V,E_b,W_b)$ :

\begin{align*} 
V = \{C\,|\,&\forall (w_i,w_j)\!\in\!C^2,\ \text{PMI}(w_i,w_j)\!>\!th_{\text{PMI}}\}, \\[-1pt]
E_c &= \{(C_i,C_j)\,|\,s(C_i,C_j)\!>\!th_c\},\quad \\[-1pt]
E_b &=\{(C_i,C_j)\,|\,s_v(\mathbf{v_{C_i}},\mathbf{v_{C_j}})\!>\!th_b\}, \\[-1pt]
&W_c(C_i,C_j) = \text{norm}\!\big(s(C_i,C_j)\big),\quad \\[-1pt]
&W_b(C_i,C_j) = \text{norm}\!\big(s_v(\mathbf{v_{C_i}},\mathbf{v_{C_j}})\big)
\end{align*}

with : 
\begin{equation}    
s_v(\mathbf{v},\mathbf{w})=
\begin{cases}
1-\dfrac{\|\mathbf{v}-\mathbf{w}\|_2}{d_{\max}}, & \text{(euclidean)}\\[4pt]
\dfrac{\mathbf{v}\cdot\mathbf{w}}{\|\mathbf{v}\|\|\mathbf{w}\|}, & \text{(cosine)}
\end{cases}
\end{equation}

\section{Comparison methods}

\subsection{Graph embedding}\label{GE}

A first idea  to compare semantic content conveyed by those two models is to start from the graph structure and embed it in a vector space. These procedures are well documented principally because it allows graph structures to be visualized while retaining relational information. In practice a graph embedding is a application that assigns each node of the graph  a vector in a $n$-dimensional space. 
This type of application gives a set of vectors, each one corresponding to a node of $G_c$ so a clique in $V$. The resulting set and its geometry can be directly compared to the emerging geometry in the vector space generated by CamemBERT. I this study we use 4 kinds on graph embeddings to enforce the comparison :
\begin{itemize}
    \item Force directed method \cite{eades1984heuristic}, an usual method to visualize graph structure in 2D. It's based on energy minimization problem in a physical system by modeling spring links.   
    \item Spectral method, based on the eigen-decomposition of the graph Laplacian. It embeds nodes by minimizing distances according to the spectrum of the Laplacian, revealing low-dimensional structures that preserve global connectivity patterns.   
    \item Isomap method \cite{tenenbaum2000global}, a technique that preserves geodesic distances between nodes on the manifold induced by the graph. It use spectral decomposition to recover the intrinsic geometry of the structure.    
    \item Node2Vec \cite{grover2016node2vec}, a random walk method that encode proximity in the emergent metric by analysing relation between nodes in random walks within the graph. It is a learning method that usually need many dimensions.  
\end{itemize}

\subsection{Comparison between graphs structure}

Another way to compare the two models is to start from the vector space generated by the CamemBERT-based procedure and to construct a graph $G_b$ whose edges reflect the similarity between vectors.
The comparison method 
applied to the structures of the two graphs $G_c$ and $G_b$ uses
%using 
graph theory tools as clustering or Breath-first Search (BFS) subgraphs for example.

Clustering enables to highlight lexical fields in the corpus using precise, controllable structural criteria. This is because cliques are grouped together by construction between cliques used globally in the same contexts. Thus, by structurally analyzing the groupings within the emerging geometries, we are able to identify areas of meaning to a greater or lesser extent, which will also serve as a criterion for the quality of the models. 
Here, we rely in particular on the Infomap method \cite{rosvall2011multilevel}. This algorithm detects communities by modeling the flow of information along random walks on the graph. The principle is to minimize the description length of a random walker’s trajectory using information theory. In other words, the algorithm seeks a partition of the graph that allows an optimal compression of movements, grouping together nodes that are frequently visited together. This approach is particularly well suited to our context, as it emphasizes the structural coherence of semantic regions and thus form zone of meaning within the global topology of structures.

The advantage of these comparison methods is that they allow to switch between well-documented 
two ways of representing the semantic geometry.

\section{Results}
In order to carry out these comparisons between the two types of models, we distinguish between the issues related to the scales highlighted by each type of comparison. We begin by applying the concept of graph embedding to examine the arrangements of meaning at the local level. Next, we attempt to compare the geometries that emerge at the global level.

\subsection{Local structures}
\subsubsection{BFS subtrees} %of $G_c$}
Let's begin by analyzing what emerges from comparing the relevance of local structures generated by both models. Each model defines, for every clique, a specific semantic neighborhood — discrete in the case of $G_c$, and continuous in the embedding space. To evaluate how these two geometries align locally, we extract subgraphs from $G_c$ using BFS trees of limited depth around a given root clique $C_i$. Each BFS subtree thus represents the local semantic environment of that clique, as defined by the discrete co-occurrence structure. So we  compare in a first place spatial distribution in the space generated by the $G_c$ embedding and spatial distribution in the vector space generated by CamemBERT using BFS subtrees of $G_c$ and vector spaces similarities. 

\begin{figure}
    \centering
    \includegraphics[width=1\linewidth]{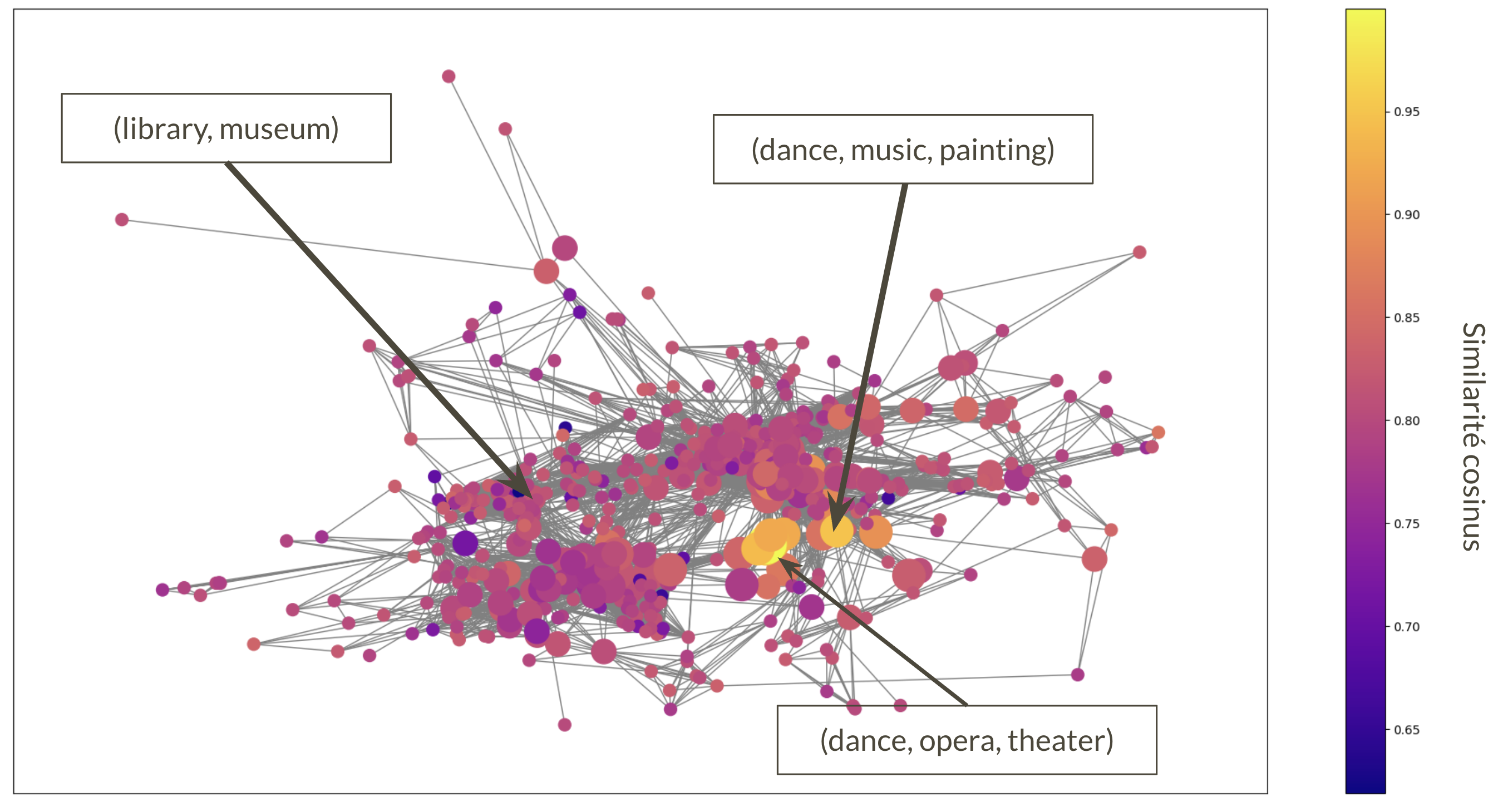}
    \caption{BFS subgraph (depth=3) of $G_c$ starting from the clique $C = $ (\textit{danse, opéra, théâtre}), (dance, opera, theater)) embedded using the force-directed method. Each node is colored according to the cosine similarity between the clique of that node and $C$ in the space generated by the CamemBERT model. The size of the nodes is inversely proportional to their distance from the clique $C$ in the BFS path. Note that labels are originally in French but translate here.
}
    \label{fig:proptheatre}
\end{figure}

Figure \ref{fig:proptheatre} shows us an example of how the local geometry of the two models is organized. The graph model  shows a decrease in semantic coherence (due to the metric distance within the embedding) over a distance consistent with the meaning we give to the semantic proximity between two cliques. We can highlight, for example, the gradation of semantic proximity that emerges between the cliques (\textit{danse, opéra, théâtre}) (dance, opera, theater) and (\textit{danse, musique, peinture}), (dance, music, painting) and between (\textit{danse, opéra, théatre}),  (dance, opera, theater) and (\textit{bibliothèque, musée}), (library, museum). Indeed, we would expect this decrease in proximity. For the geometry within the continuous model, we can see from the cosine similarities that the semantic distinction via distance is very quickly stifled. After about ten cliques, the meaning is averaged out. 

Thus, very locally, the structures generated by the models coincide. However, the models diverge in terms of their semantic description after a dozen cliques on this example. The discrete model predicts a longer progression than the graph model, which averages proximities more quickly.

It is important to note that in this example, we are able to embed the structure of $G_c$ in two dimensions because it is an example taken from a BFS with few nodes. Thus, even if the embedding comes from the complex embedding of $G_c$ (and is most likely impossible to embed “correctly” in two dimensions), we do not see any overlap with other lexical fields.

\subsubsection{Local metric correlation analysis}
To move beyond the purely visual or two-dimensional comparison of neighborhoods, we propose a quantitative assessment of the coherence between the two semantic spaces. For each node within a BFS-extracted subgraph of $G_c$, we compute two complementary measures relative to the root clique $C$ : (1) the metric distance to $C$ in the graph embedding space, and (2) the cosine similarity to $C$ in the CamemBERT space. Each node is then represented as a point in a two-dimensional scatter plot, with its graph-based distance on the $x$-axis and its BERT-based similarity on the $y$-axis. This representation allows us to evaluate the correlation between the two metrics — that is, whether the semantic proximities encoded by the discrete co-occurrence structure are preserved in the continuous embedding space.

\begin{figure}
    \centering
    \includegraphics[width=1\linewidth]{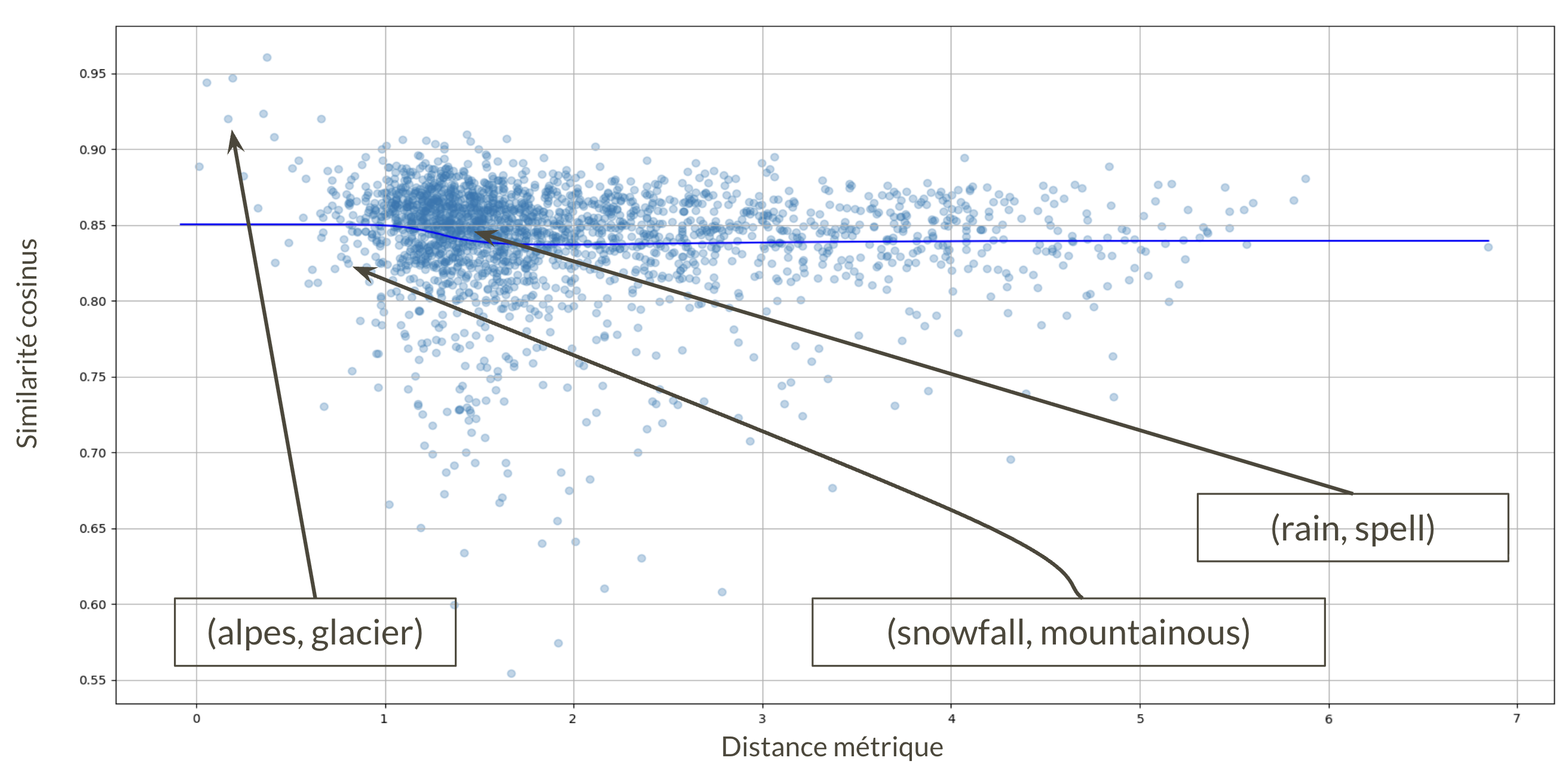}
    \caption{Correlation between metric distances in the graph embedding space (x-axis) and cosine similarity in the space generated by CamemBERT (y-axis). The starting clique is (\textit{banquise , fondre}), (ice floe, melt) and the embedding is of type Node2Vec in dimension 50 of $G_c$ constructed at $th_\text{PMI}=9$ and $p = 0.1\%$. For the readability of the local aspect of the comparison, only nodes from the BFS path of depth 6 ($\sim 1100$ cliques) are represented. The correlation coefficient approximately -0.137. Note that labels are originally in French but translate here.}
    \label{fig:propN2V}
\end{figure}

In this framework, a negative correlation would indicate that the two geometries are globally consistent: nodes that are close in the graph embedding (small $x$) should correspond to high similarity in the BERT space (large $y$), while nodes farther apart in the graph should gradually exhibit lower similarity. Conversely, the absence of such correlation would reveal a breakdown of structural coherence between the two models. 

The results exhibit a clear decay of correlation as one moves away from the root clique  $C$. In the immediate neighborhood (first BFS shell), CamemBERT tends to agree with the graph structure: the closest co-occurrence cliques remain semantically close in the embedding space. However, this correspondence quickly vanishes beyond a small radius — typically after ten to fifteen nodes as describe the correlation coefficient $r=-0.137$ by it's small size. The scatter plots become diffuse, and the correlation between graph-based distance and embedding-based similarity collapses toward zero. This pattern is illustrated in Figure~\ref{fig:propN2V}, where the local consistency around the root clique (\textit{banquise, fondre}) (ice floe, melt) is visible for only a limited set of related cliques, such as (\textit{Alpes, glacier}), (Alpes, glacier). Beyond this local range, however, semantic coherence fades: distant cliques like (\textit{enneigement, montagne}), (snowfall, mountain), although structurally connected in $G_c$, show no particular similarity in CamemBERT’s representation.

Those observations reveals a structural limitation of large language model embeddings such as CamemBERT, while they capture local semantic regularities efficiently, they fail to reproduce the gradual and progressive organization of meaning that emerges from discrete co-occurrence graphs. The continuous embedding space tends to smooth and flatten semantic distinctions, yielding dense local neighborhoods but no coherent sense of large-scale organization by this averaging effect. In contrast, $G_c$ induces a layered and gradated geometry of lexical proximity, where transitions between fields of meaning occur progressively through overlapping cliques. This gradual property is largely absent from the continuous model, which instead favors isotropic distributions driven by statistical regularities rather than explicit relational structures.

\subsection{Global structure comparison}
We now aim to compare the global organization of meaning. 
 A key finding from the previous section
suggests that, for the graph model, the semantic structure can extend gradually beyond local and specific neighborhoods. Here, we attempt
to shed further light on this effect of overall architecture.

\subsubsection{Embeddings repartition}

\begin{figure*}[t]
    \centering
    \includegraphics[width=0.84\textwidth]{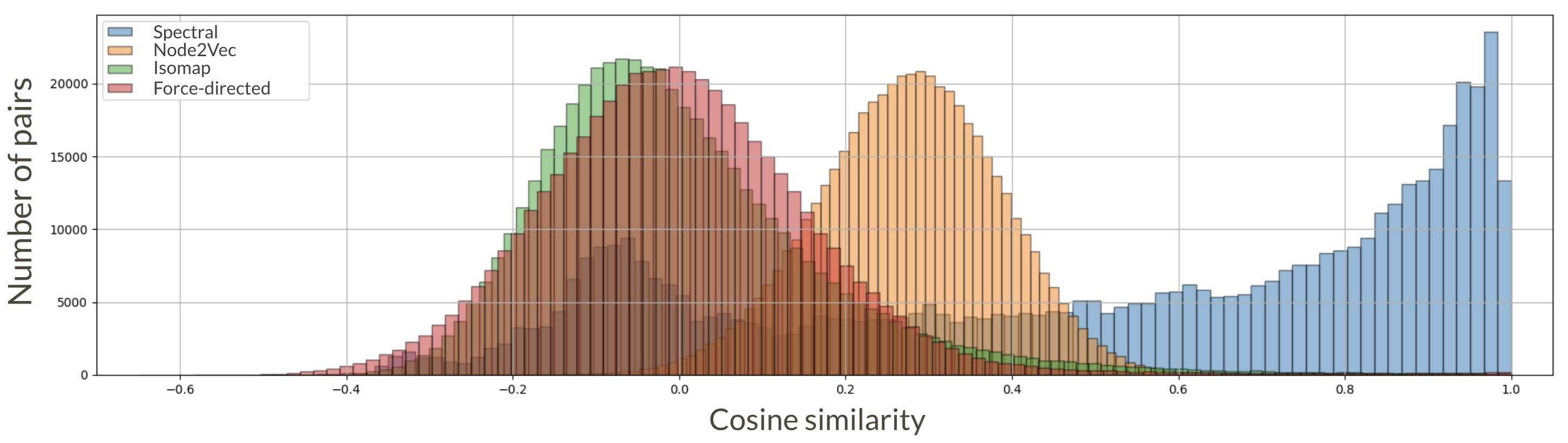}
    \caption{Distribution of cosine similarities between vectors of the four types of embeddings of $G_c$ (constructed at $th_\text{PMI}=9,\; p=0.001$) in $30$-dimensional space.}
    \label{fig:distribfull}
\end{figure*}

\quad A first step in gaining an understanding of the overall organization is to examine the distribution of vectors in the embedding spaces. The occupation of space by the embeddings provides information about the precision and convergence of the algorithms involved. \\
\quad As for the embedding of $G_c$, analyzing the vector distribution  allow us to use dimension as a convergence 
parameter. Conceptually, we expect there to be fewer and fewer cliques as proximity increases (in the sense that there are fewer ``paraphrastic'' cliques than semantically uncorrelated cliques). This observation is reflected in the embedding space as a decreasing cosine similarity distribution as the cosine tends towards~$1$. We also expect this distribution to be centered at 0, which would indicate that all directions are being exploited. Thus, the expected bell curve in this distribution would be a marker of convergence. This can be seen in Figure~\ref{fig:distribfull}, which shows the distributions for the different types of embeddings of $G_c$ in $30$ dimensional space. One can see a bell curve centered at $0$ for the Isomap and Force-Directed algorithms (referred to as FD, in the following text), indicating that they have already converged. One can see that the Node2Vec algorithm does not cover the entire space, and that the spectral algorithm has not converged when space dimension equals $30$.\\
\quad It is important to note that these distributions do not provide information about the quality of the semantic information carried by the graph $G_c$, but only give an initial measure of the quality of the embedding, i.e., whether the graph embedding, at a given dimension, can reveal its geometry or whether the implementation is correct. These distributions are a simple verification of the embeddings rather than a measure of semantic content. One example is an FD-type embedding which, due to its implementation,  have a reasonable curve even in 2D but unsatisfactory semantic content.\\

\begin{figure*}[t]
    \centering
    \includegraphics[width=0.84\linewidth]{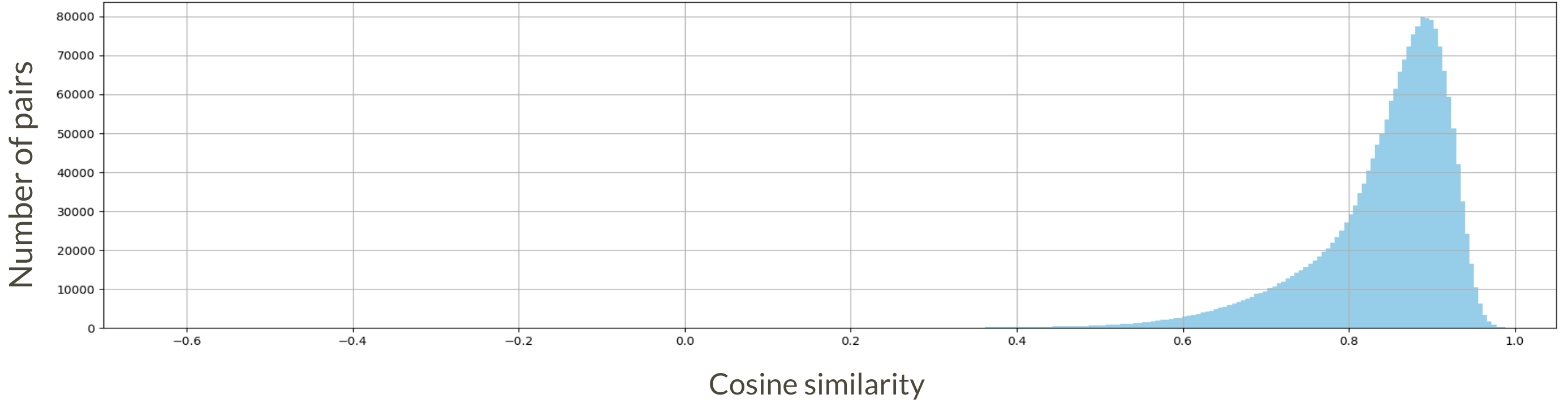}
    \caption{Distribution of cosine similarities between pairs of embeddings of cliques from the CamemBERT model.}
    \label{fig:distribbert}
\end{figure*}

\quad If we now look at the distribution of cosine similarities across all vectors generated by CamemBERT in dimension $768$, shown in Figure \ref{fig:distribbert}, one can observe the phenomenon described above. The curve is “centered” around $0.9$, which implies the existence of a cone in the vector space where all the vectors of the embedding are concentrated. This problem, documented in particular for learning models, is known as the Curse of High Dimensionality \cite{zhang2025curse}. The addition of dimensions makes the space so vast that the data is highly dispersed, and so the distances between points lose their discriminating power and exponentially more data is needed to learn effectively. It then becomes difficult for the loss function to effectively enforce the use of all dimensions. One can also note a fairly short tail to the right of the peak, which seems to reflect a rapid loss of semantic coherence. 

\quad These spatial organization measures thus provide us with some clues for understanding the convergence of semantic spaces constructions, even if it is difficult to draw any conclusions about semantic content from these analyses. To do this, we now attempt to compare the graph structures in order to characterize the global distribution from a semantic point of view.  

\begin{figure*}[t]
    \centering
    \includegraphics[width=0.9\linewidth]{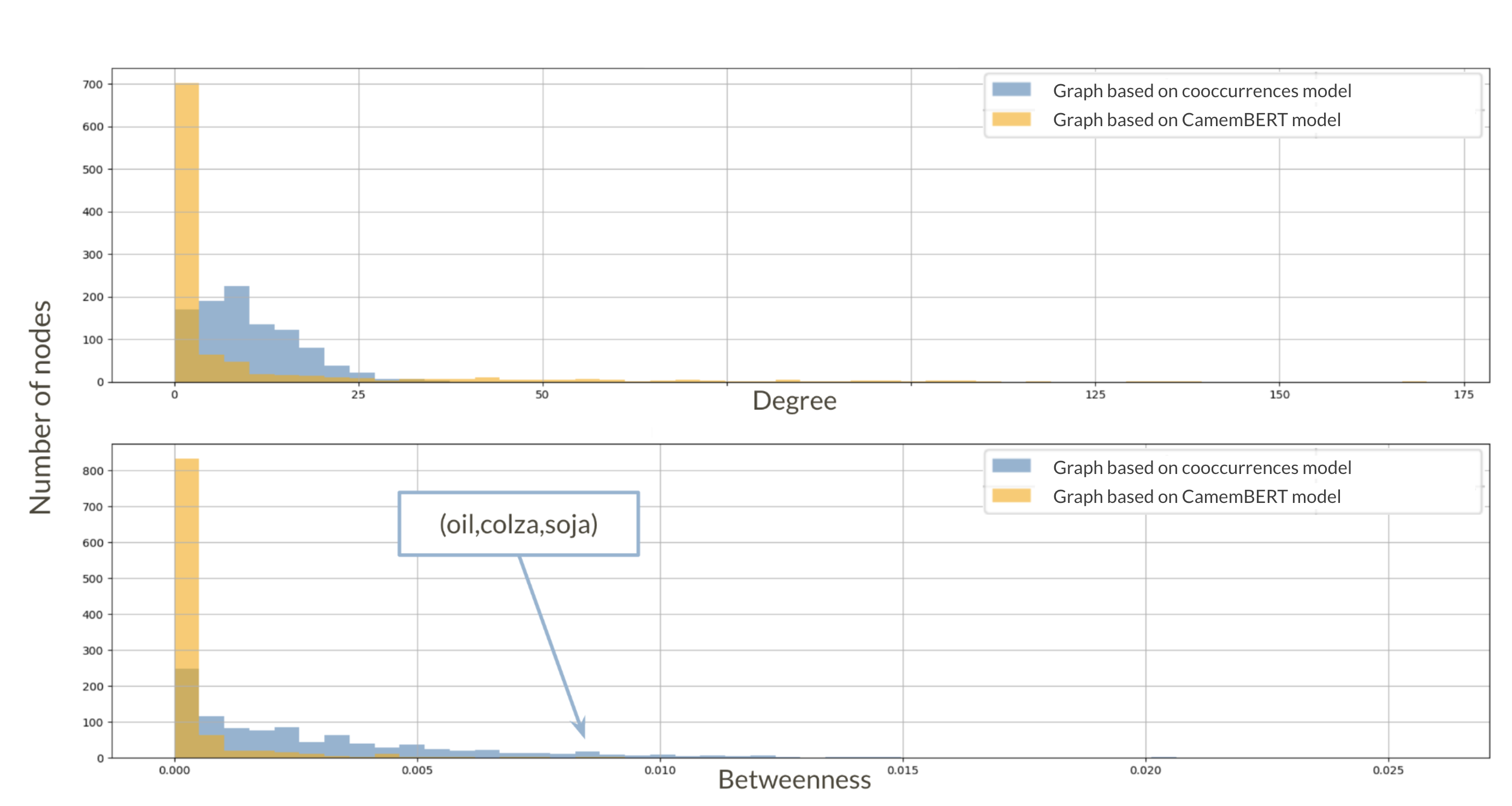}
    \caption{$G_c$ (blue) and $G_b$ (orange)  degree (top) and betweenness centrality (bottom) distributions.}
    \label{fig:btwdeg}
\end{figure*}

\subsubsection{Graph structures comparison}
In order to compare the structure  of graphs $G_b$ et $G_c$,  degree and betweenness centrality distributions were calculated for all nodes (Figures \ref{fig:btwdeg}). 
These distributions
reveal two characteristics of the structure of the   vector-based graph  $G_b$ (in orange): a few high-degree nodes connected to a 
significant portion of the graph’s nodes, and many low-degree ($< 4$) nodes that are therefore relatively isolated from the rest of the graph.
The same distributions for clique graph $G_c$ (in blue) show less variation: the peak of the degree histogram, located in the $8-12$ range, is much lower than the peak of the degree histogram for $G_b$. This indicates a smaller number of isolated or sparsely connected nodes.  Furthermore, the maximum degree, which is $25$, is much lower than that of $G_b$, which is $100$. Therefore, unlike $G_b$, $G_c$  does not contain a subset of hyper-connected nodes.  Betweenness centrality distributions, meanwhile, reveals a structural divergence regarding role nodes  within these two graphs. On the one hand, $G_b$ shows an aggregate  of betweenness centralities around low values, indicating that no node is structurally significant for paths within this graph. On the other hand, $G_c$ exhibits a wider range of betweenness centrality values, highlighting the existence of cliques that are significant for paths through the graph. These cliques with high betweenness values serve as bridges between lexical fields. For example, the clique (\textit{huile,colza,soja}) (oil, rapeseed, soy) has a betweenness centrality of $0.08$ and links, on the one hand, issues related to ecology and, on the other hand, those related to agriculture. These same centrality values are absent within $G_b$.

Based on the values of these two indicators, the graph $G_c$ has a more  homogeneous structure in terms of the distribution of links between nodes than the graph $G_b$, which contains a few central nodes with high transition power and many peripheral nodes.
This disparity should have an impact on the classification and semantic consistency of the clusters. We will demonstrate this.

\subsubsection{Clustering on semantic graphs}
\begin{figure*}[t]
    \centering
    \begin{minipage}{0.48\textwidth}
        \centering
        \includegraphics[width=\linewidth]{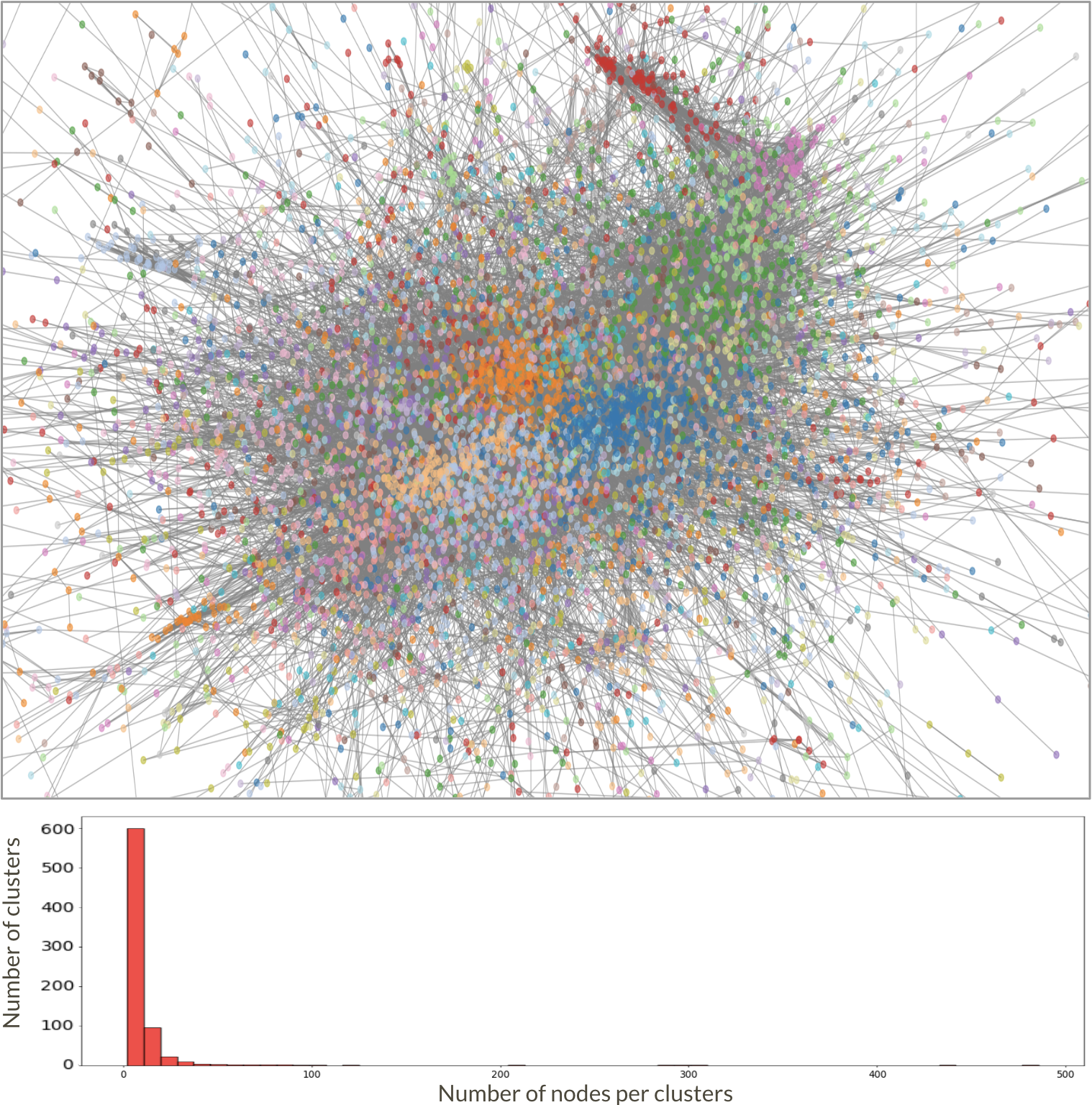}
    \end{minipage}
    \hfill
    \begin{minipage}{0.48\textwidth}
        \centering
        \includegraphics[width=\linewidth]{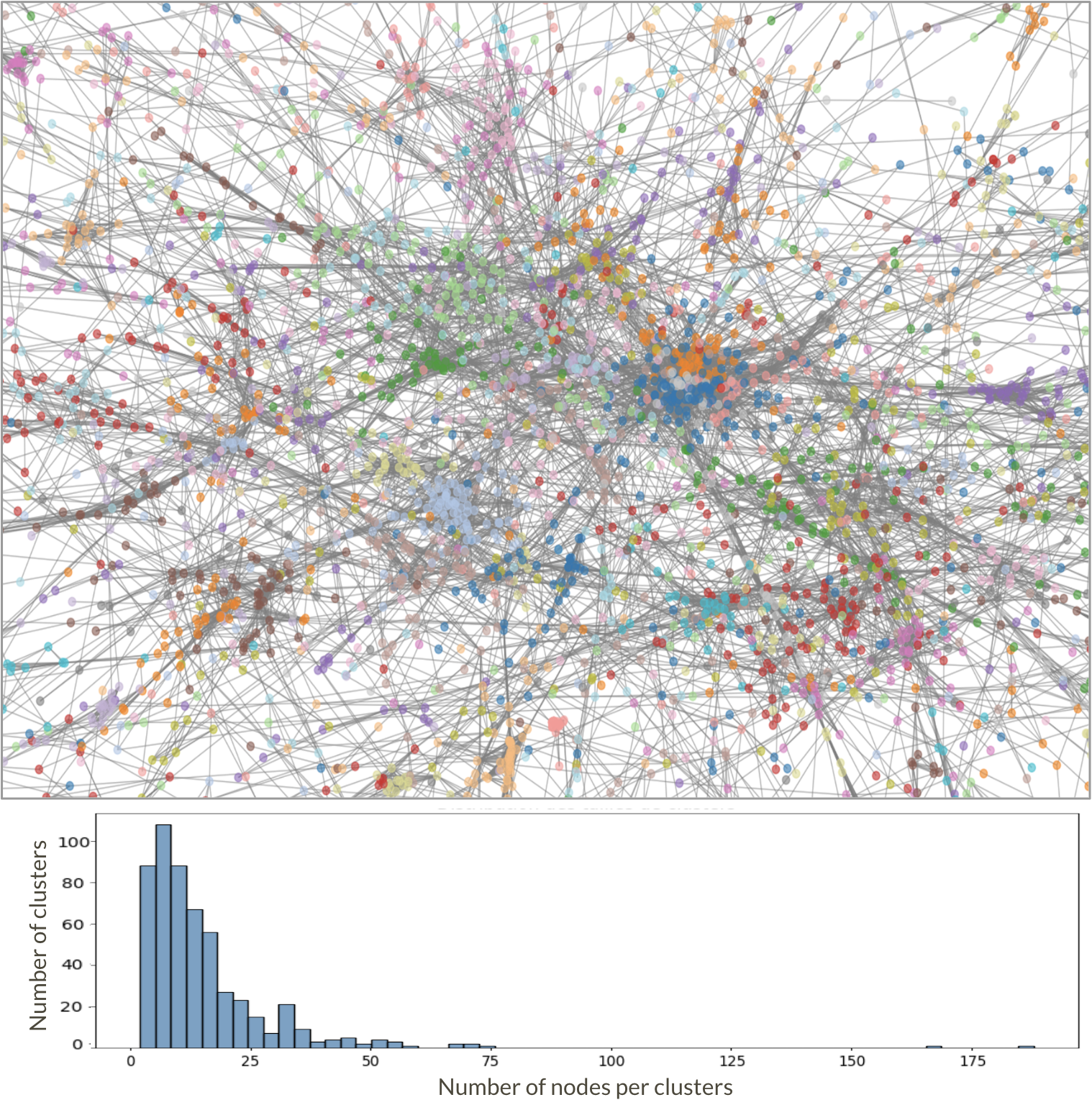}
    \end{minipage}
    \caption{Upper, $G_b$ (left) and $G_c$ (right) colored according to the Infomap partition of each graph. Graphs are constructed at $th_\text{PMI}=9$, $p=0.05\%$ and $s_\textbf{v}=s_\text{eucl}$, FD-type embedding. Lower, the respective distribution of cluster sizes in the associated Infomap partition.} 
    \label{fig:clusterim}
\end{figure*}

A community detection and graph partitioning  was
applied using the Infomap algorithm. Figure \ref{fig:clusterim}  shows the results for each
of the graphs $G_c$  and $G_b$, with the clusters colored.
Overlaid on each partition is a histogram
of cluster sizes. We
observe strong heterogeneity in $G_b$ : few
large clusters and many small
ones. The histogram for $G_c$  is flatter, with
a maximum size ($181$) much smaller than that of
$G_b$ ($480$), a peak that is not at zero, and a
steeper slope. In short, the distribution of
cluster sizes in $G_c$  is more homogeneous than
that of $G_b$.

Beyond these measures, which highlight
significant structural differences, it is important
to gain an understanding of the semantic coherence
within the clusters for each of the two graphs.
Table \ref{tab:clusterim} illustrates this aspect. It presents
cliques randomly selected from a few
of the first clusters in the  $G_b$ and $G_c$ partitions.
We observe stronger semantic coherence
between cliques within the same cluster for $G_c$ than for
$G_b$. For example, in the  largest cluster
of $G_b$, the pair of cliques
[(wind
power, energy
sector, nuclear
power), (noisy,
conversation, music)] have little thematic connection. In
contrast, based on this same random selection, one can
observe good semantic coherence starting within the
largest clusters in $G_c$.

%\begin{figure}[t]
\begin{table*}[!t]
\centering
\begin{tabular}[width=0.84\textwidth]{p{1.4cm}|p{6.5cm}|p{6.5cm}}%{|p{0.5cm}|p{3cm}|p{3cm}|}
\hline
\textbf{Cluster~\#} & \textbf{Cliques extracted from  the cluster ($G_c$)} & \textbf{Cliques extracted from  the cluster ($G_b$)} \\
\hline
1 & (\textit{cuir, foie, fourrure}) (leather, liver, fur), (\textit{cruauté, corrida, abbatoir}) (cruelty, bullfighting, slaughterhouse) & (\textit{ozone, trou}) (ozone, hole), (\textit{bonbon, bouffe, soda}) (candy, (junk)-food, soda) \\
\hline
2 & (\textit{glue, piége, chasse}) (glue, trap, hunt), (\textit{cartouche, munition, plomb}) (cartridge,  ammunition , lead shot) & (\textit{éolien, filière, nucléaire}) (wind power,  energy sector, nuclear power),  (\textit{bruyante, conversation, musique}) (noisy, conversation, music) \\
\hline
10 & (\textit{dermatologue, introuvable}) (dermatologist, unavailable), (\textit{lentille, ophtalmologue, conventionné}) (contact lenses, ophthalmologist, Medicare) & (\textit{baccalauréat, BEP}) (high school diploma, vocational diploma), (\textit{cdd, intérim}) (fixed-term contract, temporary work) \\
\hline
50 & (\textit{agnostique, penseur}) (agnostic, philosopher), (\textit{bouddhiste, célébration}) (Buddhist, celebration) & (\textit{amazonien, brésil, soja}) (Amazon, Brazil, soy), (\textit{biocarburant, colza}) (biofuel, rapeseed) \\
\hline
100 & (\textit{sécheresse, période}) (drought, period), (\textit{météo, innondation, température}) (weather, flooding, temperature) & (\textit{cannabis, récréatif, légalisation}) (cannabis, recreational, legalization), (\textit{prostitution, drogue}) (prostitution, drugs) \\
\hline
\end{tabular}
\caption{Randomly selected cliques extracted from Infomap-derived clusters (graphs $G_c$ and $G_b$)  constructed at $p=0.1\%$. The first column lists the cluster indices in descending order of size.}
\label{tab:clusterim}
\end{table*}

\section{Graph structure and embedding dimensionality}
\begin{figure*}[t]
    \centering
    \includegraphics[width=0.95\linewidth]{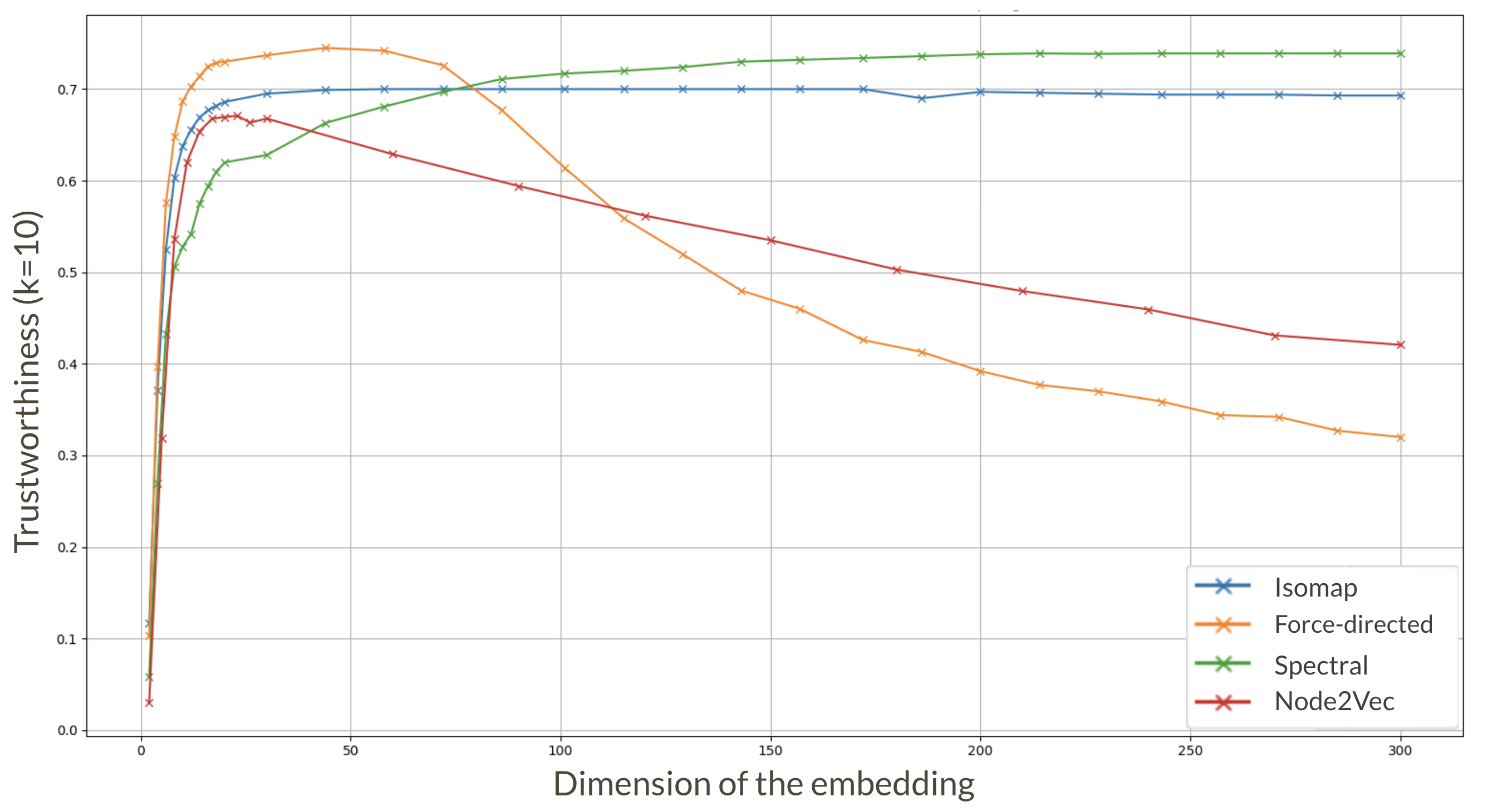}
 \caption{Evolution of trustworthiness at $k=10$ on the Infomap partitioning as a function of the dimension of the embedding of $G_c$ at $p=0.1\%$ across the four types of embeddings.}
    \label{fig:trust}
\end{figure*}

Language models are known for their
high dimensionality. By varying the dimension of the embedding space using the
methods described in Section \ref{GE}, we sought to determine the number of dimensions required for a
“good” representation of the calculated clusters.
Figure \ref{fig:trust} illustrates trustworthiness\footnote{
The trustworthiness  measure $Tw$ is the average of the number of the $k$ nearest neighbors of $e$ in the embedding space that belong to the same cluster as $e$. It ranges from $0$ to $1$; a value close to $1$ indicates that the local structure of the graph predicted by the clustering is well preserved in the embedded space.} $Tw$ depending on the dimension of the embedding. 
\begin{equation*}
    Tw_k(E, P) =
    \frac{1}{n} \sum_{i=1}^{n} 
    \frac{1}{k} \sum_{j \in \mathcal{N}_k(e_i)} 
    \mathbf{1}_{[P(e_i) = P(e_j)]}
\end{equation*}
where $E$ is the set of the $n$ vectors of the embedding, $P$ is the partition, and $\mathcal{N}_k(e_i)$ denotes the $k$ nearest neighbors of $e_i$ in the vector space.
We observe that the curves converge rapidly to a maximum value at $> 0.7$ for a dimension  $<100$ for three of the methods. This shows that the embedding space of the graph is consistent with the classification while maintaining a reasonable dimensionality.
% On remarque que les courbes convergent rapidement vers une valeur maximale à $> 0.7x$ et une dimension $<100$ pour 3 des méthodes. Ceci montre que l'espace plongé du graphe est cohérent avec la classification tout en ayant une dimensionnalité raisonnable.

\section{Conclusion}
We have developed a method for comparing the geometry of semantic spaces and the structure of the graphs induced by a language model and a regular co-occurrence clique-based model.  This comparison was conducted using the same corpus of texts. 
The results highlight a consistency in the local semantic similarities between the two types of models.
They also highlight significant differences in geometry and global structure.
Based on this comparison, the graph structure and the geometry of the embeddings generated by the co-occurrence cliques appear to demonstrate a better gradation of semantic distances, a more evenly distributed use of embedding space, and a more homogeneous graph structure in terms of degree, betweenness, and cluster size.
An assessment of the semantic coherence of the clusters also supports the discrete model. Finally, the various embeddings tested show good classification performance starting at a few dozen dimensions.

These findings call for further research. While language model training and architectures provide a highly effective solution for all generation tasks that primarily require context-local performance, could they benefit from a better representation of the overall structure of semantic spaces? Would this approach make it possible to reduce the size of the effective dimensions? 
If these assumptions were to prove true, they would result not only in quantitative and qualitative gains but also in an increase in the amount of required resources.

\section{Limits}
This study was conducted for a single language (French), a unique corpus of texts, and a sole language model (CamenBERT).  Its extension to other languages, other corpora, and different models remains to be done.

\bibliography{custom}

@article{barbaresi2023simplemma,
  title={Simplemma},
  author={Barbaresi, Adrien},
  journal={Zenodo},
  year={2023}
}

@inproceedings{ji2003lexical,
  title={Lexical knowledge representation with contextonyms},
  author={Ji, Hyungsuk and Ploux, Sabine and Wehrli, Eric},
  booktitle={Proceedings of Machine Translation Summit IX: Papers},
  year={2003}
}

@article{liu2019roberta,
  title={Roberta: A robustly optimized bert pretraining approach},
  author={Liu, Yinhan and Ott, Myle and Goyal, Naman and Du, Jingfei and Joshi, Mandar and Chen, Danqi and Levy, Omer and Lewis, Mike and Zettlemoyer, Luke and Stoyanov, Veselin},
  journal={arXiv preprint arXiv:1907.11692},
  year={2019}
}

@inproceedings{martin2020camembert,
  title={CamemBERT: a tasty French language model},
  author={Martin, Louis and Muller, Benjamin and Suarez, Pedro Ortiz and Dupont, Yoann and Romary, Laurent and de La Clergerie, {\'E}ric Villemonte and Seddah, Djam{\'e} and Sagot, Beno{\^\i}t},
  booktitle={Proceedings of the 58th annual meeting of the association for computational linguistics},
  pages={7203--7219},
  year={2020}
}

@article{zhang2025curse,
  title={Curse of High Dimensionality Issue in Transformer for Long-context Modeling},
  author={Zhang, Shuhai and You, Zeng and Chen, Yaofo and Wen, Zhiquan and Wang, Qianyue and Qiu, Zhijie and Li, Yuanqing and Tan, Mingkui},
  journal={arXiv preprint arXiv:2505.22107},
  year={2025}
}

@article{rosvall2011multilevel,
  title={Multilevel compression of random walks on networks reveals hierarchical organization in large integrated systems},
  author={Rosvall, Martin and Bergstrom, Carl T},
  journal={PloS one},
  volume={6},
  number={4},
  pages={e18209},
  year={2011},
  publisher={Public Library of Science San Francisco, USA}
}

@inproceedings{grover2016node2vec,
  title={node2vec: Scalable feature learning for networks},
  author={Grover, Aditya and Leskovec, Jure},
  booktitle={Proceedings of the 22nd ACM SIGKDD international conference on Knowledge discovery and data mining},
  pages={855--864},
  year={2016}
}

@article{tenenbaum2000global,
  title={A global geometric framework for nonlinear dimensionality reduction},
  author={Tenenbaum, Joshua B and Silva, Vin de and Langford, John C},
  journal={science},
  volume={290},
  number={5500},
  pages={2319--2323},
  year={2000},
  publisher={American Association for the Advancement of Science}
}

@article{eades1984heuristic,
  title={A heuristic for graph drawing},
  author={Eades, Peter},
  journal={Congressus numerantium},
  volume={42},
  number={11},
  pages={149--160},
  year={1984}
}

\end{document}